\documentclass[11pt]{article}
\usepackage{coling2020}
\usepackage{times}
\usepackage{url}
\usepackage{latexsym}
\usepackage{mathtools}
\usepackage{graphicx}
\usepackage{subcaption}
\usepackage[]{todonotes}
\usepackage{dsfont}
\usepackage{svg}
\usepackage{booktabs}

\colingfinalcopy % Uncomment this line for the final submission

\title{Improving Conversational Question Answering Systems \\ after Deployment using Feedback-Weighted Learning }
\author{Jon Ander Campos\textsuperscript{\rm 1}, Kyunghyun Cho\textsuperscript{\rm 2}, Arantxa  Otegi\textsuperscript{\rm 1},\\ \textbf{Aitor Soroa\textsuperscript{\rm 1},  
 Gorka Azkune\textsuperscript{\rm 1},  Eneko Agirre\textsuperscript{\rm 1}} \\ 
\textsuperscript{\rm 1}University of the Basque Country (UPV/EHU)\\
\textsuperscript{\rm 2}New York University (NYU)\\
\textsuperscript{\rm 1} \texttt{\{jonander.campos, arantza.otegi, a.soroa,} \\
\texttt{ gorka.azkune, e.agirre\}@ehu.eus}, \textsuperscript{\rm 2}\texttt{kyunghyun.cho@nyu.edu}}

\date{}

\begin{document}
\maketitle
\begin{abstract}
%Feedback weighted learning for ConvQA in LLL
The interaction of conversational systems with users poses an exciting opportunity for improving them after deployment, but little evidence has been provided of its feasibility. In most applications, users are not able to provide the correct answer to the system, but they are able to provide binary (correct, incorrect) feedback. In this paper we propose feedback-weighted learning based on importance sampling to improve upon an initial supervised system using binary user feedback. We perform simulated experiments on document classification (for development) and Conversational Question Answering datasets like QuAC and DoQA, where binary user feedback is derived from gold annotations. The results show that our method is able to improve over the initial supervised system, getting close to a fully-supervised system that has access to the same labeled examples in in-domain experiments (QuAC), and even matching in out-of-domain experiments (DoQA).  Our work opens the prospect to exploit interactions with real users and improve conversational systems after deployment.      
\end{abstract}

\section{Introduction}

%Conversational systems are usually built using manual rules, supervised machine learning or a combination of both. Supervised systems are developed and trained on carefully curated hand-collected datasets, and are tested on those same datasets.  
In Conversational Question Answering (CQA) systems, the user makes a set of interrelated questions to the system, which extracts the answers from reference text \cite{choi2018quac}. These systems are trained on datasets of human-human dialogues collected using Wizard-of-Oz techniques, where two crowdsourcers are paired at random to emulate the questioner and the answerer. Several projects have shown that it is possible to train effective systems using such datasets. For instance, QuAC includes question and answers about popular people in Wikipedia \cite{choi2018quac}, and DoQA includes question-answer conversations on cooking, movies and travel FAQs \cite{campos-etal-2020-doqa}. Building such datasets comes at a cost, which limits the widespread use of conversational systems built using supervised learning. 

The fact that conversational systems interact naturally with users poses an exciting opportunity to improve them after deployment. Given enough training data, a company can deploy a basic conversational system, enough to be accepted and used by users. Once the system is deployed, the interaction with users and their feedback can be used to improve the system. %\todo{add a brief summary of (limitations of) related work here: requirement of user providing correct answer (not realistic in all cases), lack of comparison to supervised systems, chit-chat conversations}
%\todo{User telling the right answer: This is a stronger assumption than ours, as in our case, we only require that the teacher recognizes correct and incorrect answers. }

\begin{figure}[t]
  \centering
  \includegraphics[width=\textwidth]{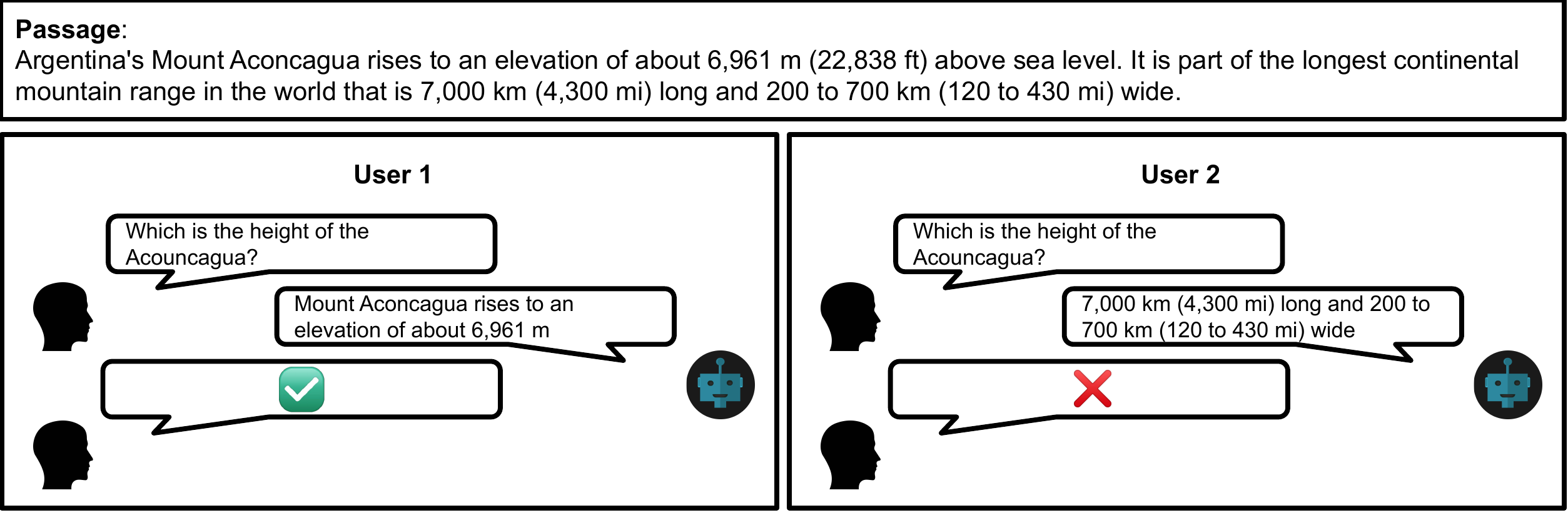}
  \caption{Example of the CQA task where at a point in the conversation the user 1 gives positive feedback to the system and user 2 gives a negative one due to the received incorrect answer.}
  \label{fig:task}
\end{figure}

In this work we focus on the case where a CQA system trained off-line is deployed and receives explicit binary (correct, incorrect) feedback from users. An example of this task can be seen in Figure \ref{fig:task} where at a point in the conversation two different users give binary feedback to the system according to the correctness of the received answer. Assuming a large number of interactions, we can safely ignore examples for which no feedback is received. We propose feedback-weighted learning (FWL) based on importance sampling as the technique to improve the initial supervised system using only binary feedback from users.

In our experiments user feedback is simulated, and the correct/incorrect feedback is extracted from the gold standard. That is, if the system output matches the gold standard output then it is deemed correct, otherwise it is taken to be incorrect.  
In order to develop and test feedback-weighted learning we perform  initial experiments on  document classification. The results show that the model improved by the proposed algorithm performs comparably to the fully supervised model that is fine-tuned with true labels rather than binary feedback. Those experiments are also used to check the impact of hyperparameters like the weight of the feedback and the balance between exploitation and exploration, which shows that our method is not particularly sensitive to the values of those hyperparameters. 

Regarding CQA, we use the best hyperparameters from the earlier experiment on document classification, and conduct experiments using several domains in CQA including datasets like QuAC and DoQA.
Our method always improves over the initial supervised system. In the in-domain experiments (QuAC) our method is close to the fully supervised model which is fine-tuned with true labels rather than binary feedback, and in the out-of-domain experiments (DoQA) our method matches it. The out-of-domain results are particularly exciting, as they are related to the case where a CQA system trained off-line in one domain could be deployed in another domain, letting the users improve it via their partial feedback by interacting with the system. Our experiments reveal that the proposed approach is robust to the choice of the system architecture, as we experimented with both multi-layer perceptron and pre-trained transformer. %Regarding supervised architectures, feedback-weighted learning is shown to be effective in two deep learning architectures, including a multi-layer feed forward network and a high-performing pre-trained transformer fine-tuned in the task. 

%Our work does the following contributions:
%\begin{itemize}
%\item We present a novel method based on importance sampling which is able %improve the results of two widely used deep learning architectures using %partial feedback only. 
%\item The results from document classification show that the proposed method improves over the initial supervised system, matching the performance of a fully supervised system which uses the true labels of deployments examples. %Results show that feedback reaches the performance of a supervised  system trained with labels => it is feasible to improve a NN using  user feedback with the same amount of “information” as the supervised method.
%\item The results on several in-domain and out-of-domain CQA experiments show that the proposed method improves over the initial supervised system in all cases, matching a supervised approach in out-of-domain experiments.
%\item The analysis shows that the method is robust to hyperparameter choice.
%\end{itemize}

The main contribution of our work is a novel method based on importance sampling, feedback-weighted learning, which improves the results of two widely used deep learning architectures using partial feedback only. Experimental results from document classification show that feedback-weighted learning improves over the initial supervised system, matching the performance of a fully supervised system which uses true labels. In-domain and out-of-domain CQA experiments show that the proposed method improves over the initial supervised system in all cases, matching a fully supervised system in out-of-domain experiments. 
This work opens the prospect to exploit interactions with real users and improve conversational systems after deployment. All the code and dataset splits are made publicly available \footnote{\url{https://github.com/jjacampos/FeedbackWeightedLearning}}.     

%\item Motivation: enpresak S0, deployment, nola hobetu S0 erabiltzaileei erantzun zuzenak eskatu gabe (bakarrik feedback BAI/EZ)? Aukeratzen dugu S0rako arkitektura neuronal superbisatu standard batzuk (MLP, pre-trained transformer), eta hori hobetzen saiatzen gara.

%\item Google Award-etik recuperatu daiteke zerbait?

%\begin{itemize}
%    \item Task Definition//Learning after deployment -> Ataza formalizatu eta feedback weighted learning %azaldu?
%    \item Ataza -> S0 sistema bat entrenatu modu superbisatuan offline eta ondoren contextual bandit multiclass setting erabiliz hobetu 
%    \item hemen sartuko zen gure soluzioa -> Importance sampling)
%\end{itemize}

%Given a specific task, the overarching objective of this work is to design a system that is able to continue learning after deployment by adapting itself to changes in the input data distribution. 

%Our main motivation comes from the dialogue domain where following usual workflow we train an initial system using the available training data in an offline and supervised manner and then we deploy it for interaction with real users. Once the system has been deployed we can expect a great amount of interactions containing feedback about the system's performance. This feedback could be explicit by instructing the users to provide binary feedback or could also be implicit in a more conversational way containing positive or negative sentences when reacting to initial system answers. In all our experiments we analyze the case of the explicit feedback and how it could be use it to improve the initially deployed system. 

\section{Related Work on Conversational Question Answering}
\label{sec:sota}

CQA research builds on reading comprehension. In reading comprehension the system has to answer questions about a certain passage of text in order to show that it understands the passage. There are two main methods: the \textit{extractive} method, in which the answer is selected as a contiguous span in the reference passage, and the \textit{abstractive} method, in which the answer text is generated. Many datasets \cite{rajpurkar2016squad,rajpurkar2018know,dunn2017searchqa,kovcisky2018narrativeqa,trischler2016newsqa,bajaj2016ms} and systems  have been proposed to address this task, where the \textit{extractive} scenario has drawn special attention \cite{wang2016machine,seo2016bidirectional}. Lately, with the incursion of large pre-trained language models as BERT \cite{devlin2019bert}, XLNet \cite{yang2019xlnet} and their relatives, the state of the art has been dominated by systems that use the representations obtained with these pre-trained language models. The systems learn answer pointer networks that consist of two classifiers, one for spotting the start token of the answer span and another for spotting the end token of the answer span. In reading comprehension, the questions are individual and isolated, that is, they do not have any dialogue structure. 

Due to the increasing interest on modelling the conversational structure behind user questions, several CQA datasets where questions and answers are interrelated have been created following the Wizard-of-Oz technique. Among all the datasets we can highlight QuAC \cite{choi2018quac}, CoQA \cite{reddy2019coqa} and DoQA \cite{campos-etal-2020-doqa}. While the first two datasets cover more formal domains as Wikipedia articles and literature, the latter covers different domains extracted from online forums as StackExchange. Contextual versions of the previously mentioned reading comprehension models have successfully modelled the conversational structure in those datasets \cite{qu2019attentive,qu2019bert,ohsugi2019simple,ju2019technical}.

\section{Importance Sampling for Learning After Deployment}
%\begin{itemize}
%    \item Task Definition//Learning after deployment -> Ataza formalizatu eta feedback weighted learning azaldu?
%    \item Ataza -> S0 sistema bat entrenatu modu superbisatuan offline eta ondoren contextual bandit multiclass setting erabiliz hobetu 
%    \item hemen sartuko zen gure soluzioa -> Importance sampling)
%\end{itemize}

In our learning after deployment scenario we start by training an initial $S_0$ system in an off-line and supervised way. This first system follows the traditional workflow where we have access to limited supervised training and development data. Then, we take the best performing system on the development data and deploy it to serve user queries. In this deployment phase, every time a user makes a query $x$, the system generates an answer $y$ and the user gives binary feedback to it. Over time, the system generates different answers $y_{i1}, y_{i2}, ..., y_{in}$ and receives feedback for each item $x_i$ . We assume a sufficient amount of user interactions, and as such we ignore any query-answer pair for which the user did not provide feedback. After the system has been deployed for a while, we collect for each question the answers provided by the system, and the respective user feedback. 

We consider a CQA system implemented using two classifiers predicting the start and end tokens respectively. This allows us to consider each classifier independently and describe the process of learning after deployment for a single classifier. We propose to use feedback-weighted learning, which is based on self-normalized importance sampling, in order to generate the system answers.

\subsection{Feedback-Weighted Learning}
\label{sec:feedb-weight-learn}

In this section, we describe a novel algorithm for updating a classifier trained off-line on-the-fly based on user feedback alone. We start by defining the true distribution $p^*(y|x)$  over $C$ classes given an input $x$. This distribution is constructed to reflect binary user feedback $\left\{ -\beta, \beta\right\}$:

\begin{align*}
p^*(y|x) \propto
\begin{cases}
\exp(\beta),&\text{if $y$ is correct} \\
\exp(-\beta),&\text{if $y$ is incorrect}
\end{cases}
\end{align*}
In words, the correctness of each class is reflected in the magnitude of the probability assigned to the class which is proportional to the user feedback. The hyperparameter $\beta$ controls the weight of the feedback.

The goal of the proposed algorithm is to minimize 
the KL divergence from $p^*$ to the classifier's predictive distribution $q(y|x; \theta)$ w.r.t. the parameters $\theta$, where
\begin{equation}
\label{eq:kl}
\text{KL}(p^*\|q) = -\sum_{y} p^*(y|x) \log q(y|x; \theta) + \mathcal{H}(p^*).
\end{equation}
Exact minimization of this objective is however intractable due to the lack of access to the true distribution $p^*$. We can instead query the unnormalized $p^*$ given the input $x$ and a candidate class $y$. 

We thus resort to self-normalized importance sampling with the following proposal distribution:
\begin{equation}
\hat{q}(y|x) = \lambda q(y|x; \theta) + (1-\lambda) \mathcal{U}(y),
\end{equation} 
where $\mathcal{U}(y)$ is a uniform distribution over $y$ and smooths out the potentially peaky predictive distribution $q$. We can control this smoothness, which trades off exploration and exploitation, by controlling the mixing coefficient $\lambda$~\cite{hoi2018online}.

With this proposal distribution, we derive the following objective function for feedback-weighted learning, starting from Eq.~\eqref{eq:kl}:
\begin{align}
\text{KL}(p^*\|q) - \underbrace{\mathcal{H}(p^*)}_{\text{const. w.r.t. } \theta} =& -\sum_{y} \hat{q}(y|x) \underbrace{\frac{p^*(y|x)}{\hat{q}(y|x)}}_{= w(y^k)} \log q(y|x; \theta) 
\nonumber\\
\approx&
-\frac{1}{K} \sum_{k=1}^K \frac{\omega(y^k)}{\sum\limits_{k=1}^K\omega{(y^k)}} \log q( y^k | x;\theta),
\end{align}
where $K$ is the total number of user feedback received. 

The importance weight $\omega(y^k)$ is computed as 
\begin{equation}
    \log \omega(y^k) = \underbrace{\beta \mathds{1}(y^k=y^*)}_{=\text{feedback}} - \log \hat{q}(y|x),
\end{equation}
where $y^*$ is the (unknown) true class, and 
\begin{align*}
    \mathds{1}(\alpha) = 
    \begin{cases}
    1, \text{ if $\alpha$ is true} \\
    -1, \text{ if $\alpha$ is false}
    \end{cases}
\end{align*}
In other words, the importance weight reflects the ratio between the user feedback and the model's confidence in each sampled prediction $y^k$. We hence call this algorithm {\it feedback-weighted learning}.

\subsection{Related Work on Lifelong Learning}

Continual or lifelong learning is defined as a system's ability to continually learn over time by accommodating new knowledge while keeping previously learned experiences \cite{parisi2019continual}. Within this framework of lifelong learning, we particularly focus on building a system that adapts to changes in the data distribution after deployment \cite{agirre2019framing}.

There have been efforts for learning actively from dialogue during deployment. The question answering (QA) setting was explored in \newcite{weston2016dialog} and  \newcite{li2017dialogue}, where they analyzed a variety of learning strategies for different dialogue tasks with diverse types of feedback. In these studies they also touch on \textit{forward prediction}, which uses explicit user correction. This idea was later applied to chit-chat systems \cite{hancock2019learning}. These works relied on users explicitly providing  the correct answer. This strong assumption was relaxed in \newcite{weston2016dialog}, where the user provides binary feedback on correct and incorrect answers in a synthetic question answering task \cite{weston2015towards}. Our work also uses binary feedback and tests it in more realistic CQA datasets.

In a similar online setup to ours, ~\newcite{liu2018customized} explored contextual multi-armed bandits for dialogue response selection using a customized version of Thompson sampling. In this work they use the Ubuntu Dialogue Corpus \cite{lowe2015ubuntu} for user simulation. In the case of task-oriented dialogue systems, \newcite{liu2018dialogue} propose a hybrid learning method with supervised pre-training and further improvement using human teaching and feedback. For the human teaching case they use imitation learning with explicit corrections done by an expert. After that, they resort to reinforcement learning for further improvement thanks to long term rewards defined by task completion.  
 
\section{Experiments}
\label{sec:experiments}

In this section we present the experiments with feedback-weighted learning (FWL). In the experiments we first build a supervised system ($S_0$), and then we simulate a deployment phase by letting $S_0$ answer user queries and receiving their feedback. User feedback is derived from a manually annotated deployment set, which is obtained by splitting the training set. We refer to the set used for training $S_0$ as a {\it training set} and the other partition of the original training set as a {\it deployment set} in the rest of the paper.  

We consider the following systems and baselines:
\begin{itemize}
\item $S_0$: the original supervised system trained on the training dataset only. We consider this system a baseline.
\item $S_0$\emph{ + FWL}: $S_0$ is fine-tuned with FWL using examples and partial feedback from the deployment set. 
\item $S_0$\emph{ + supervised}: we first train $S_0$ as above, and then continue its training using examples from the deployment set using the true labels instead of binary feedback. This is thus a fine-tuned system that has full access to the true data.
\item \emph{Fully supervised}: a supervised system trained from scratch using the union of the training and deployment sets.% We take this as an upper-bound system.
\end{itemize}

Although our main objective is to develop a lifelong learning system for CQA, we also perform experiments on document classification, as a way to assess the robustness of the proposed method when applied to different neural architectures and tasks. Moreover, these experiments are used to develop the system and check the impact of hyperparameters, so that the best hyperparameters from document classification are used in the CQA experiments. 

\subsection{Document Classification}
\label{sec:class}

% In this section we present the experiments on document classification. Although our main objective is to develop a LLL system for ConversationalQA, we also test our method in a document classification dataset with the aim of testing the robustness of the proposed method when applied to different neural architectures and tasks.

The model for document classification is a simple multi layer perceptron (MLP) with a single hidden layer. The input to the MLP is a document vector, calculated as the average of the GloVE vectors \cite{pennington2014glove} of all the words in the document. The dimension of the embeddings is set to 300, and the hidden layer has 200 hidden units.

Experiments are performed on the DBPedia Classes dataset,\footnote{
\url{https://www.kaggle.com/danofer/dbpedia-classes}
} 
which contains hierarchical categories of $342,748$ Wikipedia articles. Each article is categorized at three levels into $9$, $70$ and $219$ categories respectively. We use the latter setting with $219$ classes in our experiments. The dataset comes with a standard train, development and test splits. We kept the development and test sets untouched, but we split the training part further, creating a training set and a deployment set with the $10\%$ and $90\%$ of the original training examples, respectively. These percentages are motivated on real scenarios where the initial amount of training data is usually limited and expensive to obtain, but during deployment it could be easier to collect more data in a cheaper way. In the deployment phase we consider the feedback to be positive when the class assigned by the system is the same as the gold class in the deployment set, and negative otherwise. 

Regarding the experimental setting, the $S_0$ system is built on the train split using cross entropy loss. For the $S_0$ \emph{+ FWL} system we perform hyperparameter exploration of $\lambda \in [0.5, 1.0]$ and $\beta \in [1,85]$ using Bayesian optimization~\cite{NIPS2012_4522}. The hyperparameter values that performed best in the original development set after one epoch are selected, which corresponds to  $\lambda=0.97$ and $\beta=76$. We sample class predictions 3 times for each example, based on our preliminary experiments, and train $S_0$\emph{ + FWL} a maximum of $50$ epochs. Given $N$ the amount of training examples and $K$ the amount of samples, in this article we will use \emph{epoch} to mean $N \times K$ feedback requests. See Section \ref{sec:Discussion} for a further discussion on sample efficiency in FWL.  

% \paragraph{Settings} 
% We experiment with different...
% \begin{itemize}
%     \item $S_0$: For the training of the $S_0$ system we use supervised learning with traditional Cross Entropy loss.
%     \item $S_0$ + FWL deployment: We apply Feedback Weighted Learning (FWL) using the deployment examples to the $S_0$ system with partial feedback. Hyperparameters Lambda, Beta, epoch, sample. We set sample to 10 based on intuition and some exploratory experiments and a max. of 30 epochs with early stopping (5 wait). We performed a hyperparameter exploration of lambda and beta using bayesian ..... XXX, where lambda ranged from X to X and beta from X to X. We only one epoch. The best resulting hyperparameters are Lambda 0.99, Beta 55.
%     \item $S_0$ + supervised deployment
%     \item Fully supervised: train + deployment together
% \end{itemize}

Table \ref{tab:doc_class} shows that the simple MLP architecture performs well on this task, even when only the $10\%$ of training examples are used. Still, $S_0$\emph{ + FWL} is able to improve the performance of $S_0$ by $5$ points, and it is close to both supervised systems. These results validate the effectiveness of FWL as a way of improving an initial supervised system using binary feedback only.

\begin{table}[t]
\centering
\begin{tabular}{llr}
% \hline
Systems                   & F1             \\ 
                %   \hline
                \toprule
$S_0$              & 86.51          \\ 
$S_0$  + FWL       & 91.59  (+5.0) \\ 
$S_0$ + supervised & 91.89  (+5.3) \\ 
Fully supervised   & 92.04 (+5.5)  \\ 
% \hline
\end{tabular}
\caption{Results as F1 on document classification. Number in parenthesis for difference with respect to $S_0$. FWL continues learning over $S_0$ using only binary feedback, and the result is close to the supervised systems.}
\label{tab:doc_class}
\end{table}

\subsection{Conversational Question Answering}
\label{sec:convqa}

In the CQA experiments we fine-tune a pretrained  BERT~\cite{devlin2019bert} for QA. Given a query and a passage that contains the answer, the pretrained BERT is fine-tuned to predict the start and end indexes of the answer span. This approach has shown strong performance on QA datasets such as SQuAD \cite{rajpurkar2016squad}. In our experiments we use the base uncased model of BERT with the maximum context size of $384$ and a batch size of $12$, using default values for the rest of the hyperparameters.

We experiment with the following settings:

\begin{itemize}
\item In-domain vs. out-of-domain. We experiment with two different scenarios, based on the mismatch between training and deployment distributions. In the first scenario the domain is the same for both  training and deployment phases, whereas in the out-of-domain scenario the domains differ. %, as we used QuAC for training the $S_0$ system, as well as during deployment.
%  In the second scenario, the domain of the deployment phase is different from the one used for training the $S_0$ system. %The system is trained using QuAC, but we use DoQA for development, deployment and test.
\item Without vs. with dialogue history. In order to take into account the multi-turn feature of a dialogue, we prepend the previous question and its corresponding answer to the input. Following usual practice \cite{qu2019bert}, we consider only the previous interaction (one questions and one answer).
%  BERT does not model dialogue history. In order to take into account the multi-turn feature of a dialogue, we append the previous question and answer to the current question for introducing the dialogue history.
\end{itemize}
In the in-domain experiments we use QuAC~ \cite{choi2018quac} for both building the initial $S_0$ system and during the deployment phase. QuAC is a conversational dataset extracted from the Wikipedia using the Wizard of Oz method and crowdsourcing. In the out-of-domain scenario QuAC is used for building $S_0$, but the deployment phase is done with DoQA~\cite{campos-etal-2020-doqa}, which is a conversational dataset based on FAQs and contains dialogues from three different domains (cooking, travel and movies). 

Similarly to document classification, we split the original training parts of QuAC into training and deployment splits containing $10\%$ and $90\%$ of the training dialogues, respectively. We consider the feedback to be positive whenever the answer span predicted by the system matches the gold span exactly, and negative otherwise. Because the QuAC test split is unavailable, we report results in the development split. 

With respect to the system settings used for the experiments, we set the $\lambda$ and $\beta$ hyperparameters of the $S_0$ system based on their best values from document classification ($\lambda = 0.97$ and $\beta = 76$). Given that the CQA system contains two classifiers and the number of classes is often larger than in document classification task, we use a larger number of samples, $50$ in this task.

% \begin{table}[t]
% \centering
% \begin{tabular}{llr}
% \hline
%                    & F1            \\ \hline
% \textit{without dialogue history} && \\ \hline
% $S_0$              & 46.01         \\ 
% $S_0$ + FWL        & 50.30 (+4.3)  \\ 
% $S_0$ + supervised & 53.11 (+7.3)  \\ 
% Fully supervised   & 53.87 (+7.9) \\ \hline
% \textit{with dialogue history}   &&  \\ \hline
% $S_0$              & 48.90          \\ 
% $S_0$ + FWL        & 53.83 (+4.9)  \\ 
% $S_0$ + supervised & 55.09 (+6.2)  \\ 
% Fully supervised   & 55.03 (+6.1) \\ \hline
% \end{tabular}
% \caption{Results on conversational QA using QuAC dataset both for training and deployment. F1 accuracy results on QuAC development split.}
% \label{tab:convqa-quac}
% \end{table}

\begin{table}[t]
\centering
\begin{tabular}{lll}
% \hline
  Systems                 & no history   & dialogue history \\
\toprule
  $S_0$              & 46.76        &       49.03     \\
  $S_0$ + FWL        & 49.33 (+2.6) &  53.07 (+4.0)     \\
  $S_0$ + supervised &  53.66 (+6.9) &  55.10 (+6.1)     \\
 Fully supervised   &  54.50 (+7.7) &  55.40 (+6.5)     \\ 
%   \hline
\end{tabular}
\caption{Results of in-domain experiments using QuAC dataset both for training and deployment, with and without dialogue history. F1 accuracy results on QuAC development split. Number in parenthesis for difference with respect to $S_0$. FWL is able to improve over $S_0$ which validates its usefulness in CQA.}
\label{tab:convqa-quac}
\end{table}

Table \ref{tab:convqa-quac} shows the results on the in-domain experiments on the QuAC dataset. For each system we report the results after 3 epochs following \newcite{qu2019bert}. The results follow the trend observed in the document classification setting. Applying FWL after $S_0$ improves the results by $2.6$ and $4$ points, which confirms that FWL is a valid technique to continue fine-tuning a CQA system after deployment. Using dialogue history improves the results of all systems by almost $3$ points, stressing the importance of modeling history on CQA systems. However, the main conclusions remain unchanged. $S_0$\emph{ + FWL} still outperforms $S_0$ using only binary feedback, and is close to the supervised systems.

Table \ref{tab:convqa-doqa} shows the results when $S_0$ is trained on QuAC, and the user feedback is simulated using examples from DoQA. In these experiments we perform model selection on the development split of DoQA (which corresponds to the cooking domain) and report the results on the test datasets comprised of the cooking, travel and movies. We report only experiments using dialogue history, as this setting is more realistic for a CQA system. $S_0$\emph{ + FWL} outperforms $S_0$ across all the domains. $S_0$ \emph{+ FWL} furthermore matches the $S_0$ \emph{+ supervised} system
in the movies and travel domains, although it fails to do so in the cooking domain. The fully supervised system performs worse than $S_0$\emph{+ supervised} on this dataset, which we conjecture is due to the fact that QuAC contains more training examples than DoQA, with a ratio of approximately 3 to 1. This may cause the fully supervised system to be more biased towards QuAC, and thus yields worse results in DoQA. Note that in the $S_0$ \emph{+ supervised} system QuAC examples are used to train $S_0$ only, which is then fine-tuned with DoQA examples, and obtains better results overall. All in all, these results suggest that the FWL approach is robust when there is a domain shift between the training and test datasets.

\begin{table}[t]
\centering
\begin{tabular}{llll}
%\hline
Systems                             & Cooking      & Movies       & Travel        \\ \toprule
%\textit{with dialogue history} &                                             \\ \hline
$S_0$                          &    39.79     &  40.89        &  35.64     \\ 
$S_0$ + FWL                    &  49.66 (+9.9) &  47.28 (+6.4) &  47.19(+11.6) \\ 
$S_0$ + supervised             &  50.63 (+10.8) &  46.79 (+5.9)  &  47.12(+11.5) \\ 
Fully supervised               &  50.33 (+10.5) &  45.56 (+4.7)  &  46.10(+10.5)                                \\ %\hline
\end{tabular}
\caption{Results of out-of-domain experiments (with history modeling) using QuAC for training and DoQA during deployment. F1 accuracy results on DoQA test split on cooking, movies and travel domains. Number in parenthesis for difference with respect to $S_0$. FWL improves the results of $S_0$ and matches supervised results in two domains.}
\label{tab:convqa-doqa}
\end{table}

%\section{analysis}

\section{Discussion}
\label{sec:Discussion}

As shown by the experiments in document classification and CQA we are able to improve an initial supervised $S_0$ system just by using binary feedback obtained by simulating the users. In this section we perform a further analysis on several aspacts of the method.%The result is of special interest when there is a domain shift between the training data and the data received by users after deployment, which is a problem that many real systems face. 

\begin{figure}
\begin{subfigure}{.49\textwidth}
  \centering
  \includegraphics[width=1.0\linewidth]{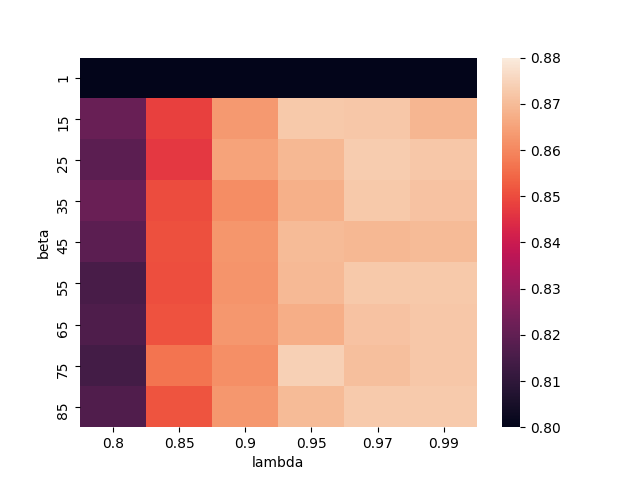}
  \caption{F1 scores obtained after one epoch and using 3 samples}
  \label{fig:first_epoch}
\end{subfigure}
\begin{subfigure}{.49\textwidth}
  \centering
  \includegraphics[width=1.0\linewidth]{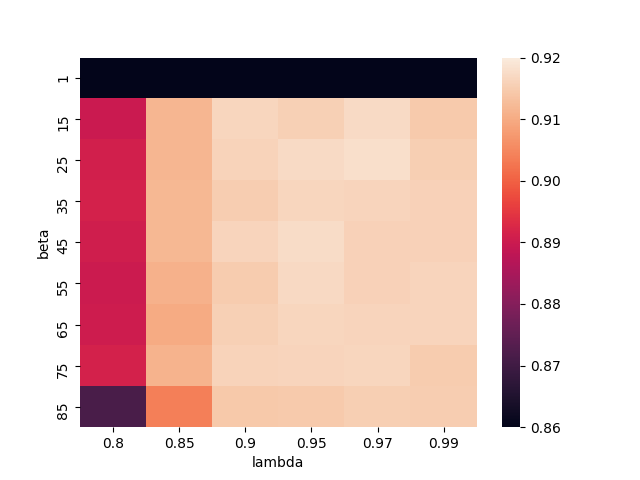}
  \caption{F1 scores obtained after 50 epochs and using 3 samples}
  \label{fig:total_train}
\end{subfigure}
\caption{Hyperparameter analysis using heatmaps on document classification showing the obtained F1 scores (lighter is better) in the development split. Similar performance is obtained with different hyperparameter pairs, showing the robustness of the method.}
\label{fig:heatmaps}
\end{figure}

\paragraph{Hyperparameters.} 
In order to show the robustness of FWL we perform several experiments in the document classification task with different values for the main hyperparameters of the method, $\lambda$ and $\beta$ (cf. Section \ref{sec:feedb-weight-learn}). The analysis shows that when using values larger than 1 for $\beta$, FWL performs similarly well for all lambdas greater than 0.8 (see Figures \ref{fig:first_epoch} and \ref{fig:total_train}). The behavior of $\lambda$ in the same Figures \ref{fig:first_epoch} and \ref{fig:total_train} reveals that large values of $\lambda$ yields best results for all beta values. In any case, the similar performance obtained with different hyperparameter combinations shows that our method is robust and not specially sensitive to small variations in the hyperparameters. %in the first epoch, but when training over 50 epochs smaller $\lambda$ perform better. Contrary to usual practice, in FWL exploitation helps at the beginning and exploration is preferred as learning goes on. This can be a consequence of starting the FWL fine-tuning from an already trained system and not from a randomly initialized model. 
%We also compared default parameters that are $\lambda = 0.8$ and $\beta = 5$ for the in-domain experiments in conversational QA and obtained an F1 score of $52.21$ for the no-history model. This result is even better than the performance obtained when using the best hyperparameters set from the document classification task $50.30$ F1 (see Section \ref{sec:convqa} for a more detailed explanation).

\begin{figure}
\begin{subfigure}{.49\textwidth}
  \centering
  \includegraphics[width=1.0\linewidth]{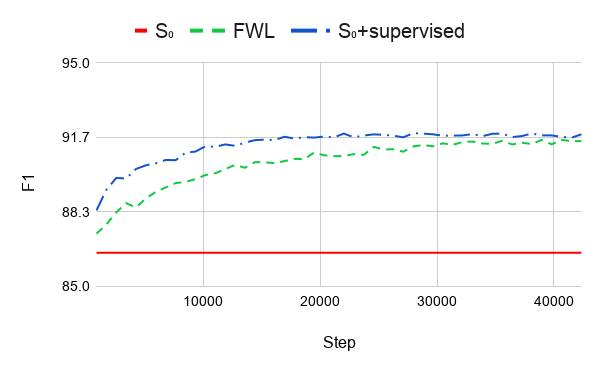}
  \caption{Document classification}
  \label{fig:doc-class}
\end{subfigure}
\begin{subfigure}{.49\textwidth}
  \centering
  \includegraphics[width=1.0\linewidth]{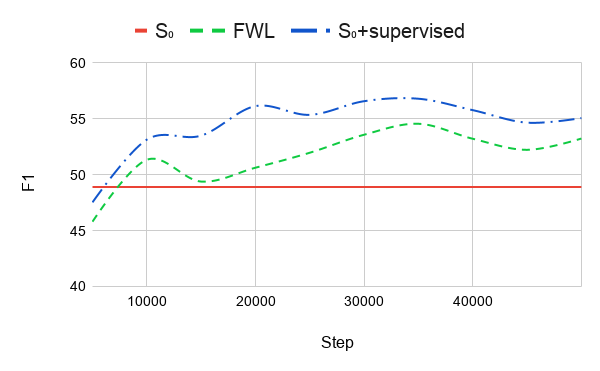}
  \caption{CQA, in-domain (QuAC dataset), with dialogue history}
  \label{fig:quac-with-history}
\end{subfigure}
\caption{Learning curves for the document classification and CQA tasks where FWL is compared to supervised learning. As the number of steps increase FWL gets closer to $S_0$ \emph{+ supervised}.}
\label{fig:quac-learning-curve}
\end{figure}

\paragraph{Learning dynamics.} 
From the learning curves in Figures \ref{fig:doc-class} and \ref{fig:quac-with-history} we see how the behavior is similar in both document classification and CQA learning tasks. In both cases the supervised systems converge faster than the FWL systems, but as the steps go on the F1 scores in the development set also converge. It is of special interest the point where FWL improves over $S_0$. In the document classification task FWL improves over $S_0$ in the first steps, and by the end of the first iteration, which comprises circa 850 steps, it already outperforms $S_0$. In CQA FWL needs more steps but the improvement over $S_0$ also happens at the beginning of the training process.

\paragraph{Sampling vs. supervised learning.} Since we treat epochs in FWL as in supervised learning, we sample new answers for each new epoch. For example, in the document classification case we end up taking $150$ samples ($50$ epochs with $3$ samples per epoch) for a total of $219$ classes. It can be argued that a dummy sampling technique covering all classes is equivalent to having the true label, and would be similar to our method in terms of sampling efficiency. However, when deploying a $S0$ system in a realistic scenario, the dummy sampling strategy would return low probability responses and could severely hamper user engagement. In contrast, our sampling method tends to return high probability answers, making it more user-friendly. In any case, each time the loss gradient is computed, FWL has information of only $3$ samples, unlike supervised learning where all classes are considered. Besides, $3$ samples per example (one epoch) are enough for FWL to improve over $S_0$ (see Figure \ref{fig:total_train}), although the best results are obtained after $50$ epochs.

\paragraph{Assumptions and limitations.}

We discuss a few assumptions we made in designing the proposed FWL. In all our experiments we simulate user feedback using supervised data, and thus the feedback is always accurate and explicit. We therefore do not consider the case where the user is unsure about the response it gave to the system, which would cause a noisy feedback that can harm the performance of the system. Moreover, as we need more than one sample for each question we would need different users making the same questions if we were to compare our method with real use-cases. Analyzing the impact of these issues and possible solutions to them is kept as an open research question for future analysis.

%Analysis of what this system implies

%\begin{itemize}

%\item DISCUSSION: Premisa batzuk/limitations: feedback-a accurate da eta esplizitua. Erabiltzaileak simulatzen ditugu etiketatutako datuetan oinarrituta. Ez gara kezkatzen erabiltzailean sortu dezakeen rechazoa. Ez gara kezkatzen sampling efficiency neurtzen, unsure ignoratzen ditu guztiz (berez, ez ditugu). Sample kopurua ez dugu mugatzen => etorkizunean sample < klase.

%\item DISCUSSION: galdera bakoitzaren hainbat sample behar dira (!), horrek esan nahi du erabiltzaileek galderak errepikatu behar dituztela. Hori gertatzeko soluzioa posibleak etorkizunerako, galderak eta feedback-ak pilatzen joatea: galdera bat egiten den bakoitzean lehenago erantzundako galderen zerrenda aurkeztea EDO sistemak erabili dezala paraphrase detection (~similarity) algoritmo bat galderak multzokatzeko, eta feedback bera erabili dezala multzo bakoitzean.

%\end{itemize}

\section{Conclusion and Future Work}

In this work we propose feedback-weighted learning that allows a supervised classifier to effectively adapt itself after deployment from partial user feedback. The experiments show that our technique is successful, in that it improves over the initial supervised system. More specifically, in document classification experiments, it matches an off-line supervised system trained with all the true labels, although it has only access to the binary feedback. More importantly, the experiments in two widely used CQA datasets, QuAC and DoQA, confirm that it is feasible to improve a CQA system after deployment.
In the DoQA experiments, the CQA system is trained off-line in one domain (Wikipedia) and then deployed in other domains, letting the users improve it via their partial feedback by interacting with the system. In this setting, the performance of our model also matches that of the fully supervised model which is fine-tuned with true labels rather than binary feedback.  Moreover,  feedback-weighted learning is shown to be effective in two deep learning architectures, including a multi-layer feed forward network and a high-performing pre-trained transformer fine-tuned in the task.

This work uses simulated feedback derived from gold standard labels. In the  future we plan to modify feedback-weighted learning to cope with noisy feedback, as well as modifying it to work with fewer samples per query. 
%has several limitations which need to be tackled in the future. On the one hand, the experiments have been performed with simulated feedback. Real user feedback poses additional challenges, including inaccurate feedback and how to collect repeated questions. 

%All the code and dataset splits developed within this work are going to be publicly available for reproducibility purposes\footnote{Url to be defined upon acceptance}.

\section*{Acknowledgments}

This research was partially supported by a Google Faculty Award, EU ERA-Net CHIST-ERA LIHLITH funded by the Agencia Estatal de Investigación (AEI, Spain) project PCIN-2017-118, the Basque Government excellence research group (IT1343-19) and the NVIDIA GPU grant program. Jon Ander Campos enjoys a doctoral grant from the Spanish MECD. Kyunghyun Cho was partly supported by Samsung Advanced Institute of Technology (Next Generation Deep Learning: from pattern recognition to AI) and Samsung Electronics (Improving Deep Learning using Latent Structure), and thanks CIFAR, eBay, NVIDIA and NAVER for their support.

\bibliographystyle{coling}
\bibliography{coling2020}

\end{document}